\documentclass{article}
\usepackage{ijcai26}

\usepackage{times}
\usepackage{soul}
\usepackage{url}
\usepackage[hidelinks]{hyperref}
\usepackage[utf8]{inputenc}
\usepackage[small]{caption}
\usepackage{graphicx}
\usepackage{amsmath}
\usepackage{amsthm}
\usepackage{booktabs}
\usepackage{algorithm}
\usepackage{algorithmic}
\usepackage[switch]{lineno}

\usepackage{bm}
\usepackage{tabularx}
\usepackage{float}

\title{Predicting The Cop Number Using Machine Learning}

\author{
Meagan Mann
\and
Christian Muise\and
Erin Meger
\affiliations
Queen's University
}

\begin{document}

\maketitle

\begin{abstract}\label{sec:abstract}
    Cops and Robbers is a pursuit evasion game played on a graph, first introduced independently by Quilliot \cite{quilliot1978jeux} and Nowakowski and Winkler \cite{NOWAKOWSKI1983235} over four decades ago. A main interest in recent the literature is identifying the cop number of graph families. The cop number of a graph, $c(G)$, is defined as the minimum number of cops required to guarantee capture of the robber. Determining the cop number is computationally difficult and exact algorithms for this are typically restricted to small graph families. This paper investigates whether classical machine learning methods and graph neural networks can accurately predict a graph's cop number from its structural properties and identify which properties most strongly influence this prediction. Of the classical machine learning models, tree-based models achieve high accuracy in prediction despite class imbalance, whereas graph neural networks achieve comparable results without explicit feature engineering. The interpretability analysis shows that the most predictive features are related to node connectivity, clustering, clique structure, and width parameters, which aligns with known theoretical results. Our findings suggest that machine learning approaches can be used in complement with existing cop number algorithms by offering scalable approximations where computation is infeasible.
\end{abstract}

\section{Introduction}\label{sec:intro}
Cops and Robbers is a pursuit-evasion game played on a graph, where $k$-cops try to capture a single robber. The cops are first placed anywhere on the graph, followed by the robber, then they each take turns making a move by either remaining on the current vertex or moving to a neighbouring vertex. Multiple cops can occupy the same vertex at any point during the game. When $k$-cops are able to successfully capture the robber by having at least one cop occupy the same vertex as the robber, we say that $G$ is $k$-copwin. The minimum such $k$ is said to be the cop number of $G$, denoted $c(G).$

Traditionally, research in this area has focused on deriving theoretical bounds for the cop number of graphs with certain properties ~\cite{https://doi.org/10.1002/jgt.23194,bonato2017characterizationsalgorithmsgeneralizedcops,Mohar_2025,gahlawat2023parameterizedanalysiscopsrobber}  \nocite{bonato2017characterizationsalgorithmsgeneralizedcops}\nocite{Mohar_2025}\nocite{gahlawat2023parameterizedanalysiscopsrobber}. Some bounds are known for certain graph families, however, many others remain unresolved. Determining the cop number is computationally difficult in general \cite{KINNERSLEY2015201}. Existing exact algorithms are typically limited to relatively small graph families as a result of the exponential growth of the underlying state space.

In this paper, we investigate whether classical Machine Learning (ML) and Graph Neural Networks (GNNs) can predict the cop number of graphs from their structural properties. Extending to larger graphs, where higher cop numbers become more common but exact computation of entire classes is infeasible, is the main motivation for applying machine learning in this domain. Using a curated dataset of graphs together with computed invariants, we develop and evaluate models that estimate the cop number of a graph and assess which features are most informative for this predictive task. In addition to predictive performance, another objective is to determine whether the informative features correspond to or refine existing theoretical bounds on the cop number known in the literature. This study aims to complement current theory on graph structure influencing cop number and identify promising future directions for further analysis. A secondary goal is for the approach to be scalable and generalizable to larger graph classes where computation is infeasible. We evaluate success by measuring both standard predictive accuracy metrics and through the interpretability of the learned models.

This paper addresses the following research questions:

\noindent\textbf{RQ1: Can the cop number of a graph be predicted accurately using classical ML models trained on handcrafted graph properties?}
Yes. Our results show that classical machine learning models, particularly tree-based methods such as Gradient Boosting, can predict the cop number of graphs up to size 13 with high accuracy using handcrafted graph properties. Even with a class imbalance these models achieve strong performance, indicating that cop number is closely related to structural properties of the graph.

\noindent\textbf{RQ2: How do graph neural networks trained on graph structure compare to classical ML models for cop number prediction?}
GNNs achieve comparable accuracy in predicting the cop number without relying on hand crafted properties. This suggests that the information relevant to cop number prediction is encoded in the graph structure and can be learned directly from graph topology.

\noindent\textbf{RQ3: Which structural graph features are the most predictive of the cop number?}
Graph properties concerning clustering, connectivity, clique size, and treewidth are the most predictive of all the handcrafted features analyzed. These feature rankings align closely with known theoretical results \cite{diestel2017graph,treewidth,clow2025copsrobberscliquecovers}\nocite{treewidth,clow2025copsrobberscliquecovers}.

\section{Related Work}\label{sec:rel}
Sun et al. \cite{sun2025graphproptraininggraphfoundation} propose a new method, GraphProp, for training Graph Foundation Models (GFMs) for strong generalization ability on graph-level tasks by pretraining them on graph invariants rather than node features. The paper shows that the graph structure itself provides much more cross-domain information compared to labeled features, and that learning structural representations enables few-shot generalization. This paper supports a core assumption of our project that structural properties of graphs can be used to predict high-level graph properties. GraphProp emphasizes learning from structure rather than relying on node features. Sun et al. focus on pretraining general-purpose graph representations, whereas our work only studies the cop number. We also do not aim to build a universal foundation model for the scope of this project, rather we pair classical feature-based ML methods and GNNs to determine which best capture structural properties relevant to bounding the cop number.

Ying et al. \cite{ying2019gnnexplainergeneratingexplanationsgraph} introduce GNNExplainer, the first framework to provide interpretable explanations of GNN predictions. Their method involves identifying the most influential subset of node features that drive the model's output to give interpretable reasoning for decisions. Interpretability is the main focus of our RQ3, which aims to determine which structural properties are most predictive of the cop number. GNNExplainer provides an approach for identifying substructures that correlate with specific cop number results. While GNNExplainer focuses on explaining black-box GNN behavior, our project aims to compare feature interpretability across classical ML models. Our goal is not only to explain, but also to test whether learned predictive output aligns with the current bounds in the literature of the cop number.

\section{Dataset}\label{sec:data}
\subsection{Populating the Dataset}
We constructed the database used in this project for research in 2024. The graphs are sourced from McKay’s Small Graphs Database\footnote{https://users.cecs.anu.edu.au/~bdm/data/graphs.html}, where all graphs are isomorphism free, meaning no duplicates. The dataset consists of all connected graphs up to 9 vertices, totalling approximately 270,000 graphs, including final cop numbers and about 40 structural features of the graphs computed using custom-built functions and NetworkX library \cite{Hagberg2008}. The graph properties computed are listed in Table \ref{tab:graph-features}. 
\begin{table}[t!]
\centering
\renewcommand{\arraystretch}{1.2}
\begin{tabular}{p{3.4cm} p{4.6cm}}
\hline
\textbf{Feature Category} & \textbf{Computed Properties} \\
\hline
Graph Size & Number of nodes, number of edges, density \\

Connectivity & Approximate node connectivity, all-pairs node connectivity, node cuts, contains bridge \\

Clique \& Independence & Clique size, $k$-clique communities, clique removal, chordal graph cliques, largest independent set size \\

Width Parameters & Treewidth, chordal graph treewidth \\

Clustering \& Local Structure & clustering coefficient, average neighbor degree \\

Distance-Based & Diameter, radius, eccentricity, average shortest path length, Wiener index \\

Domination \& Covering & Domination number, min weighted vertex cover size, min edge cover \\

Structural & Planarity, chordality, asteroidal-triple free, spectral bipartivity \\

Algorithmic & Cop number, randomized partitioning, one-exchange \\
\hline
\end{tabular}
\caption{Structural graph properties computed for each graph in the dataset}
\label{tab:graph-features}
\end{table}


The set of graphs is exhaustive, so it fully represents the structural diversity of all small connected graphs of size 2 to 9 vertices. The diversity of this exhaustive dataset is important because models trained on a narrow scope would not generalize.

\subsection{Expansion}
The dataset has been expanded to include a limited number of 1-cop-win, 2-cop-win, and 3-cop-win graphs on graphs of size 10 to 13, with some degree restrictions. These graphs are from a result in Turcotte and Yvon's paper on the minimum order of 4-cop-win graphs \cite{TURCOTTE202174}. They solved a question posed in \cite{minon10} by computationally identifying all 3-cop-win graphs on 11 vertices. Particularly, they identify 

\begin{itemize}
    \item 1 connected graph $G$ on 10 vertices
    \item 6 connected graphs $G$ on 11 vertices with $\Delta(G) \le 5$
    \item 80 connected graphs $G$ on 12 vertices with $\Delta(G) \le 4$
    \item 173 connected graphs $G$ on 12 vertices with $\Delta(G) \le 5$
    \item 1,105 connected graphs $G$ on 13 vertices with $\Delta(G) \le 4$
\end{itemize}

\noindent such that $c(G) = 3$, where $\Delta(G)$ is the maximum degree of the graph. All other connected graphs $G$ considered with these orders and maximum degrees satisfy $c(G) \le 2$ \cite{TURCOTTE202174}. 

They also identify
\begin{itemize}
    \item 457 connected graphs $G$ on 10 vertices with $\delta(G) \ge 2 \text{ and } \Delta(G) \le 3$
    \item 1,359 connected graphs $G$ on 11 vertices $\delta(G) \ge 2 \text{ and } \Delta(G) \le 3$
    \item 4,819 connected graphs $G$ on 12 vertices $\delta(G) \ge 2 \text{ and } \Delta(G) \le 3$
    \item 16,635 connected graphs $G$ on 13 vertices $\delta(G) \ge 2 \text{ and } \Delta(G) \le 3$
\end{itemize}

\noindent such that $c(G) = 1$, $c(G) = 2$, or $c(G) = 3$, where $\delta(G)$ is the minimum degree of the graph \cite{TURCOTTE202174}. Note that these graphs are all unique up to isomorphism.

The resulting class imbalance reflects an inherent structural property of small graphs rather than a limitation of data collection. In particular, graphs with cop number 3 are rare at small orders. For example, the class of graphs with 10 vertices contains over 11 million graphs, and there is only one graph with a cop number of 3: the Petersen graph \cite{minon10}. Including the exhaustive class of 10 vertices would have dramatically increased the dataset size while contributing almost no additional examples of higher cop numbers. Instead, we augment the dataset with graph families identified in prior work by Turcotte and Yvon \cite{TURCOTTE202174}, which include graphs with cop numbers ranging from 1 to 3 and 10 to 13 vertices. This adds approximately 23,000 graphs and introduces greater structural diversity in higher-cop-number instances while keeping the dataset computationally manageable. Our study is intentionally scoped such that exact cop numbers are computable from existing algorithms, allowing supervised learning. As graphs grow exponentially with size, higher cop numbers appear more frequently but exact cop number computation becomes infeasible, so we apply ML to help guide the search.

Table ~\ref{tab:dataset-breakdown} shows the current breakdown of the database by size and cop number. 

\renewcommand{\arraystretch}{1.2}
\begin{table}[t!]
    \centering
    \begin{tabularx}{\columnwidth}{cXXX}
    \hline
    \textbf{Graph Size} & \bm{$c(G)=1$} & \bm{$c(G)=2$} & \bm{$c(G)=3$} \\
        \hline
        2  & 1      & 0        & 0     \\
        3  & 1      & 0        & 0     \\
        4  & 5      & 1        & 0     \\
        5  & 16     & 5        & 0     \\
        6  & 68     & 44       & 0     \\
        7  & 403    & 450      & 0     \\
        8  & 3{,}791  & 7{,}326    & 0     \\
        9  & 65{,}561 & 195{,}519  & 0     \\
        10 & 7      & 450      & 1     \\
        11 & 12     & 1{,}341     & 6     \\
        12 & 21     & 4{,}543     & 253   \\
        13 & 35     & 15{,}495    & 1{,}105 \\
        \hline
    \end{tabularx}
    \caption{Graph size and cop number breakdown of the dataset}
    \label{tab:dataset-breakdown}
\end{table}
\renewcommand{\arraystretch}{1.0}

\textbf{Computing the Cop Number}.
All cop numbers were computed using the current fastest algorithm from Petr et al.'s ``A faster algorithm for Cops and Robbers'' \cite{PETR202211}, which we implemented and tested. Efficiency and cop numbers were validated by comparing against previously published and verified algorithm implementations such as \cite{2011}.

\section{RQ1}\label{sec:RQ1}
\subsection{Methodology}
To address RQ1, we use an expanded SQLite database containing approximately 300,000 connected graphs on 2–13 vertices, each annotated with its exact cop number and a collection of handcrafted structural features listed in Section~\ref{sec:data}. The preprocessing pipeline is fully automated. First the database is loaded, the graph ID column is then removed from the feature matrix, and the cop number column is then identified as the label. All remaining columns are transformed into a singular number. Any features that cannot be meaningfully represented as a number are automatically removed. For example, some features are represented as a list or tuple, so we extract the minimum, maximum, or average wherever we can meaningfully do so, resulting in 34 valid structural features. Since feature scales vary widely, all features are standardized using z-score normalization. If there are any missing values in the dataset then they are imputed using median values to reduce bias. The dataset is inherently highly imbalanced with cop number 1 taking up 23.58 $\%$, cop number 2 taking up 75.95$\%$, and cop number 3 taking up 0.46$\%$. To address the imbalance of the dataset, we use a 80/20 stratified split for proportional representation of each cop number class.

Each scikit-learn classifier uses the feature preprocessor (imputation, normalization, and numeric conversion) before classification. 

A Decision Tree was the first model to be implemented with  \textit{DecisionTreeClassifier} and \textit{class\_weight=``balanced"}. The balanced class weight adjusts weightings so that minority cop number classes receive proportionally higher weight during split decisions. We used a Decision Tree because it is interpretable and captures complex feature interactions \cite{sklearn-tree-doc}.

The second model is the Random Forest implemented with \textit{RandomForestClassifier}, \textit{n\_estimators=200}, \textit{class\_weight=``balanced\_subsample"}, and \textit{n\_jobs=-1}. Each tree is trained on a bootstrap subsample where a new dataset is created by sampling with replacement from the training set, and for each bootstrap sample, scikit-learn computes weights based on the sampled class distribution \cite{sklearn-random-forest-doc}. This model provides a stable nonlinear baseline.

The third model is the Histogram-based Gradient Boosting implemented with \textit{HistGradientBoosting} and \textit{max\_iter=200}. Standard gradient boosting evaluates unique features when computing splits, whereas the histogram-based version first discretizes continuous data into a fixed number of bins and reduces the computational cost from linear in the number of unique values to the number of bins \cite{sklearn-hgbt-example}.

The final model is logistic regression  implemented with \textit{LogisticRegression}, \textit{max\_iter=500}, \textit{class\_weight=``balanced"}, \textit{multi\_class=``auto"}. This provides a linear baseline to test whether the structural graph features are linearly separable about the cop number. The \textit{multi\_class} parameter controls how Logistic Regression handles more than two classes \cite{sklearn-logisticregression}.

\subsection{Results and Analysis}
Traditional machine learning models achieved extremely high predictive performance, showing that structural features alone are highly informative for determining the cop number. Empirical results on the full merged database for Decision Tree, Random Forest, Histogram-based Gradient Boosting, and Logistic Regression are detailed in Table ~\ref{tab:dec}, Table ~\ref{tab:ran}, Table ~\ref{tab:his}, Table ~\ref{tab:log}, respectively.

\renewcommand{\arraystretch}{1.2}
\begin{table}[t!]
\centering
\begin{tabularx}{\columnwidth}{lXXX}
\hline
\textbf{Metric} & \textbf{Class 1} & \textbf{Class 2} & \textbf{Class 3} \\
\hline
Precision & 0.9339 & 0.9767 & 1.0000 \\
Recall    & 0.9250 & 0.9797 & 0.9927 \\
F1-score  & 0.9294 & 0.9782 & 0.9963 \\
Support   & 13{,}985 & 45{,}035 & 273 \\
\hline
\multicolumn{4}{l}{\textbf{Overall Accuracy:} 0.9668} \\
\multicolumn{4}{l}{\textbf{Macro-F1:} 0.9680} \\
\hline
\end{tabularx}

\vspace{1em}
\textbf{Confusion Matrix:}
\[
\begin{pmatrix}
12936 & 1049 & 0 \\
916 & 44119 & 0 \\
0 & 2 & 271
\end{pmatrix}
\]
\caption{Decision Tree Results}
\label{tab:dec}
\end{table}
\renewcommand{\arraystretch}{1.0}

\renewcommand{\arraystretch}{1.2}
\begin{table}[t!]
\centering
\begin{tabularx}{\columnwidth}{lXXX}
\hline
\textbf{Metric} & \textbf{Class 1} & \textbf{Class 2} & \textbf{Class 3} \\
\hline
Precision & 0.9394 & 0.9882 & 1.0000 \\
Recall    & 0.9625 & 0.9807 & 0.9963 \\
F1-score  & 0.9508 & 0.9845 & 0.9982 \\
Support   & 13{,}985 & 45{,}035 & 273 \\
\hline
\multicolumn{4}{l}{\textbf{Overall Accuracy:} 0.9765} \\
\multicolumn{4}{l}{\textbf{Macro-F1:} 0.9778} \\
\hline
\end{tabularx}

\vspace{1em}
\textbf{Confusion Matrix:}
\[
\begin{pmatrix}
13460 & 525 & 0 \\
868 & 44167 & 0 \\
0 & 1 & 272
\end{pmatrix}
\]
\caption{Random Forest Results}
\label{tab:ran}
\end{table}
\renewcommand{\arraystretch}{1.0}

\renewcommand{\arraystretch}{1.2}
\begin{table}[t!]
\centering
\begin{tabularx}{\columnwidth}{lXXX}
\hline
\textbf{Metric} & \textbf{Class 1} & \textbf{Class 2} & \textbf{Class 3} \\
\hline
Precision & 0.9396 & 0.9901 & 1.0000 \\
Recall    & 0.9685 & 0.9807 & 0.9927 \\
F1-score  & 0.9538 & 0.9854 & 0.9963 \\
Support   & 13{,}985 & 45{,}035 & 273 \\
\hline
\multicolumn{4}{l}{\textbf{Overall Accuracy:} 0.9779} \\
\multicolumn{4}{l}{\textbf{Macro-F1:} 0.9785} \\
\hline
\end{tabularx}

\vspace{1em}
\textbf{Confusion Matrix:}
\[
\begin{pmatrix}
13545 & 440 & 0 \\
871 & 44164 & 0 \\
0 & 2 & 271
\end{pmatrix}
\]
\caption{Histogram-based Gradient Boosting Results}
\label{tab:his}
\end{table}
\renewcommand{\arraystretch}{1.0}

\renewcommand{\arraystretch}{1.2}
\begin{table}[t!]
\centering

\begin{tabularx}{\columnwidth}{lXXX}
\hline
\textbf{Metric} & \textbf{Class 1} & \textbf{Class 2} & \textbf{Class 3} \\
\hline
Precision & 0.8280 & 0.9874 & 0.9784 \\
Recall    & 0.9615 & 0.9378 & 0.9963 \\
F1-score  & 0.8898 & 0.9620 & 0.9873 \\
Support   & 13{,}985 & 45{,}035 & 273 \\
\hline
\multicolumn{4}{l}{\textbf{Overall Accuracy:} 0.9437} \\
\multicolumn{4}{l}{\textbf{Macro-F1:} 0.9464} \\
\hline
\end{tabularx}

\vspace{1em}
\textbf{Confusion Matrix:}
\[
\begin{pmatrix}
13447 & 538 & 0 \\
2793 & 42236 & 6 \\
0 & 1 & 272
\end{pmatrix}
\]
\caption{Logistic Regression Results}
\label{tab:log}
\end{table}
\renewcommand{\arraystretch}{1.0}

All four classical ML models achieved consistently high accuracy. Decision Tree got an accuracy of 0.9668 and a macro-F1 of 0.9680. It's confusion matrix shows that the model can't fully separate cop number 1 and 2. The Random Forest improved this to 0.9765 accuracy and 0.9778 macro-F1, and reducing misclassifications of class 1 by more than half relative to the single tree. The best performing model was Histogram-based Gradient Boosting, with 0.9779 accuracy and a macro-F1 of 0.9785. Logistic Regression performed worse than the tree-based models, with 0.9437 accuracy and 0.9464 macro-F1. This model achieved high recall for cop number 1, but it suffered from reduced precision and noticeably more confusion between classes 1 and 2, and is the first model to mistake cop number 2 for cop number 3. This shows the difficulty linear models face when separating nonlinear structural patterns. 

Examining the confusion matrices revealed several consistent patterns. Cop number 3 is classified almost perfectly in all cases, achieving between 99.3$\%$ and 100$\%$ recall. These results could reflect strong structural distinctiveness between these graphs. Although the small number of 3-cop-win graphs could lead to overfitting,the consistent performance across the four models suggests that these graphs genuinely form a highly separable cluster in the feature space. Most misclassifications occurred with cop numbers 1 and 2 because they make up the largest and most structurally overlapping classes. The ensemble methods using trees were able to reduce the missclassifications relative to the single Decision Tree. Random Forest and especially Histogram-based Gradient Boosting achieve both higher recall for class 1 and higher precision for class 2, demonstrating the benefit of using many nonlinear decision boundaries. These patterns suggest that the class imbalance and structural similarity between cop numbers 1 and 2, rather than noise or model limitations, are cause for the remaining error.\\

Overall, classical ML methods are able to predict the cop number of small graph families with high accuracy using a set of hand crafted graph features. The best-performing classifier was Histogram-based Gradient Boosting, which achieved a macro-F1 score of 0.9785. This indicates that the structural descriptors used effectively capture most of the variation that distinguishes graphs with different cop numbers. The errors that remain are concentrated almost entirely between cop numbers 1 and 2, the two classes that are both the most frequent and structurally the most similar. In contrast, graphs with cop number 3 form a small but highly distinctive cluster, leading all models to classify them with near perfect recall. These results demonstrate that structural graph properties encode rich and learnable signals related to pursuit–evasion difficulty, and they strongly motivate the feature-importance and interpretability analysis carried out in RQ3.

\section{RQ2}\label{sec:RQ2}
\subsection{Methodology}
For RQ2, we evaluate whether a deep learning model can learn the cop number directly from graph structure without relying on handcrafted features. We use the Graph Isomorphism Network (GIN) architecture. GIN is a highly expressive architecture designed to capture subtle structural differences between graphs using the Weisfeiler–Lehman graph isomorphism test \cite{douglas2011weisfeilerlehmanmethodgraphisomorphism}. This property makes it very well-suited for cop number prediction, where small structural variations can significantly influence outcomes.

All graphs analyzed are from the same database used in RQ1, except the hand crafted structural features are ignored for the purposes of this RQ. We convert each graph into a PyTorch Geometric Data object using degree as the only node-level feature. Node degrees are efficient to compute and correlate strongly with key structures of the graph. Learning from graph topology forces the GNN to infer all higher-level structure itself without feature engineering. Degree is standardized per graph to stabilize learning. The final architecture contains 3 GINconv layers for hierarchical structural feature extraction, BatchNorm after each layer to stabilize the distribution of node embeddings, ReLU nonlinearities for MLP expressiveness in GIN, global\_add\_pool to sum all node embeddings to create a graph embedding, and a MLP classifier to convert graph embedding to cop number class. Originally we used a 0.5 dropout, which is the default value from the PyTorch documentation, but this harmed learning due to over sparsification of embeddings \cite{pytorch_dropout}. Lowering to 0.3 improved stability and final metrics. The training setup consists of an Adam optimizer (1e-3) , 30 epochs, Batch size 16, 80/20 random train/test split, and multiclass cross-entropy loss. Evaluation metrics compare directly with RQ1, we report accuracy, macro-F1, macro-precision, macro-recall, full 3x3 confusion matrix. Macro metrics are essential because cop numbers are imbalanced. 

\subsection{Results and Analysis}
Table ~\ref{tab:gnn-results} shows the results of the GNN for cop number prediction after 30 training epochs.

\renewcommand{\arraystretch}{1.2}
\begin{table}[t!]
    \centering
    \begin{tabularx}{\columnwidth}{>{\arraybackslash}X >{\arraybackslash}X}
    \hline
    \textbf{Metric} & \textbf{Value} \\
    \hline
    Accuracy & 0.9465 \\
    Macro F1-score & 0.8774 \\
    Macro Precision & 0.9540 \\
    Macro Recall & 0.8278 \\
    \hline
    \end{tabularx}

\vspace{1em}
\textbf{Confusion Matrix:}
\[
\begin{pmatrix}
12260 & 1750 & 0 \\
1324 & 43683 & 0 \\
0 & 100 & 176
\end{pmatrix}
\]
\caption{GNN Results}
\label{tab:gnn-results}
\end{table}
\renewcommand{\arraystretch}{1.0}

Despite using only degree as a node feature, and no global handcrafted features, the GNN reaches 94.65$\%$ accuracy. This is significant because it demonstrates that cop number is largely encoded in the graph topology. In the initial interpretation we see that cop number 1 and 2 are predicted with high precision and recall. Cop number 3 is extremely rare in the dataset, yet the model successfully identifies 176 of 276 test cases. No graph of cop number 1 or 2 is ever predicted as 3, meaning no false high-cop-number errors. This is desirable because overestimating the cop number is far worse than underestimating it. As researchers continue to search for Meyniel extremal examples ~\cite{baird2013meynielsconjecturecopnumber,bonato2022meynielextremalfamiliesgraphs}\nocite{bonato2022meynielextremalfamiliesgraphs}, if we overestimate the cop number, we may falsely find a counter example to Meyniel's Conjecture stating the $c(G) = O(\sqrt n)$ \cite{HOANG1987302}. These results show that if we see a high cop number, we can trust it's reliability. However, there were 100 cases where the cop number is 3 but is incorrectly labeled as cop number 2. This is not ideal because the model is not accurately identifying high cop number instances, meaning Meyniel extremal graphs could be missed when scaling.

Overall, a GIN with degree-only features achieves 94.65$\%$ accuracy, nearly matching the classical ML pipeline in RQ1. The model extracts meaningful patterns directly from the graph topology. The network avoids overestimating the cop number, which is a desirable property for searching for graphs with high cop numbers, but still underestimates some cases of high cop number. The GNN can serve as a fast filter before running exact $k$-copwin algorithms.

Table~\ref{tab:main-results} summarizes the performance of all models. Classical methods achieve the strongest overall results, while the GNN achieves accuracy but lower macro-F1, indicating challenges in minority-class generalization.

\begin{table*}[t]
\centering

\renewcommand{\arraystretch}{1.2}
\begin{tabular}{lcccccc}
\toprule
\textbf{Model} 
& \textbf{Accuracy} 
& \textbf{Macro-F1} 
& \textbf{F1 (Class 1)} 
& \textbf{F1 (Class 2)} 
& \textbf{F1 (Class 3)} \\
\midrule
Decision Tree 
& 0.9668 & 0.9680 & 0.9294 & 0.9782 & 0.9963 \\

Random Forest 
& 0.9765 & 0.9778 & 0.9508 & 0.9845 & 0.9982 \\

HistGB 
& \textbf{0.9779} & \textbf{0.9785} & \textbf{0.9538} & \textbf{0.9854} & \textbf{0.9963} \\

Logistic Regression 
& 0.9437 & 0.9464 & 0.8898 & 0.9620 & 0.9873 \\
\midrule
GNN
& 0.9465 & 0.8774 & -- & -- & -- \\
\bottomrule
\end{tabular}
\caption{Performance comparison of classical machine learning models and a graph neural network for cop number prediction. Classical models report per-class F1-scores, while the GNN reports macro-averaged metrics only.}
\label{tab:main-results}
\end{table*}

\section{RQ3}\label{sec:RQ3}
\subsection{Methodology}
To address RQ3, we analyze the highest-performing tree-based model obtained in RQ1 with the goal of identifying which structural graph properties contribute the most to cop number prediction. We apply SHAP (SHapley Additive exPlanations) to the Random Forest classifier for interpretability, as it achieves the strongest performance among tree-based models. SHAP rankings produce importance scores of how each feature contributes to the model's predictions. For computational efficiency, SHAP values are computed on a sample of 800 training graphs using \verb|TreeExplainer| with the \verb|tree_path_dependent| setting. Since we are working with multiple classes, SHAP values are averaged to obtain a global feature ranking.

Since many graph invariants are highly correlated (e.g., edge count and density), SHAP’s independence assumptions do not strictly hold. To mitigate this, we use permutation importance and observe agreement between SHAP and permutation rankings.

\subsection{Results and Analysis}
Table ~\ref{tab:shap_results} summarizes the top 10 features ranked by mean absolute SHAP value.

\renewcommand{\arraystretch}{1.2}
\begin{table}[t!]
    \centering
    \begin{tabularx}{\columnwidth}{l r}
    \hline
    \textbf{Feature} & \textbf{Mean SHAP} \\
    \hline
    Node connectivity & 0.0924 \\
    Num of edges & 0.0631 \\
    Maximal clique size & 0.0306 \\
    Largest clique size & 0.0303 \\
    Num of nodes & 0.0294 \\
    Treewidth & 0.0226 \\
    Min weighted vertex cover & 0.0225 \\
    Chordal graph treewidth & 0.0154 \\
    Average neighbor degree\_5 & 0.0150 \\
    Max independent set size & 0.0148 \\
    \hline
    \end{tabularx}
    \caption{SHAP Results.}
    \label{tab:shap_results}
\end{table}
\renewcommand{\arraystretch}{1.0}

The most influential predictors are as follows. 

\textbf{1. Node connectivity approximation}. Node connectivity approximation has the highest mean SHAP of 0.0924, which is by far the strongest predictor. Higher vertex connectivity implies fewer articulation points and overall association with higher cop numbers. This aligns with known theory that
when graphs have stronger connectivity, the robber has a greater number of possible moves that could evade capture \cite{book}. 

\textbf{2. Number of edges}. The number of edges has a mean SHAP value of 0.0631. The edge count serves as a proxy for density. Denser graph often require more cops because the robber has more adjacent choices and cycles to exploit \cite{BonatoChiniforooshanPralat2010}.

\textbf{3- 4. Maximum and largest clique size.} The maximum clique size has a mean SHAP value of 0.0306 and largest clique size is 0.0303. Larger cliques restrict robber mobility locally but also increase overall density. This dual behavior helps differentiate cop number classes \cite{2011}.

\textbf{5. Number of nodes}. The number of nodes has a mean SHAP value of 0.0294, so graph size still remains predictive, though weaker than density or connectivity. Larger graphs tend to have a wider range of structural possibilities, but size alone does not determine cop number \cite{NOWAKOWSKI1983235}.

Overall, the SHAP results show that the model heavily relies on connectivity, density, clique structure, and treewidth related parameters. These are consistent with theoretical insights. Node connectivity is the strongest SHAP feature, and matches the classic understanding that robber evasion becomes easier in graphs with multiple disjoint paths \cite{clow2025copsrobbersgraphspath}.

To assess the robustness of SHAP-based attributions, we compute permutation importance using the trained Random Forest classifier. Permutation importance measures the decrease in classification accuracy when the values of a single feature are randomly permuted, so it quantifies the model's dependence on that feature. Table ~\ref{tab:perm_results} breaks down the top 10 features ranked by permutation importance on the Random Forest model.

\renewcommand{\arraystretch}{1.2}
\begin{table}[t!]
    \centering
    \begin{tabularx}{\columnwidth}{l r}
    \hline
    \textbf{Feature} & \textbf{Perm. Importance} \\
    \hline
    Min clustering coefficient & 0.0910 \\
    Average clustering coefficient & 0.0692 \\
    Spectral bipartivity & 0.0652 \\
    Largest clique size & 0.0250 \\
    Chordal graph & 0.0180 \\
    Graph has bridges & 0.0173 \\
    Node connectivity approximation & 0.0130 \\
    Max clustering coefficient & 0.0061 \\
    Average neighbor degree$\_0$ & 0.0048 \\
    Min all-pairs node connectivity & 0.0048 \\
    \hline
    \end{tabularx}
    \caption{Permutation Results.}
    \label{tab:perm_results}
\end{table}

The highest ranked features from permutation importance are minimum clustering coefficient and average clustering coefficient. These features measure how interconnected the graph is, aligning with the importance of dense substructures in the game dynamics \cite{gahlawat2023parameterizedanalysiscopsrobber}. Spectral bipartivity, clique size, and chordality  also ranked high reflecting their effect on robber mobility \cite{clow2025copsrobberscliquecovers,Kosowski2015}\nocite{Kosowski2015}.

Notably, several features that rank highly under SHAP, such as  number of edges, density, and treewidth are ranked lower with permutation importance. This difference is mainly due to strong correlations between the graph features as density, clique size, clustering, and treewidth encode overlapping structural information. Permutation importance tends to reduce weight on redundant features, while SHAP distributes importance across correlated features. As a result, the two methods emphasize complementary aspects of feature influence.

Taking SHAP and permutation importance together, connectivity and clustering dominate all features in predicting the cop number. Treewidth and clique structure are secondary but strong predictors, suggesting the model suggesting that they act as supporting signals for prediction. The model's most influential features match graph-theoretic expectations \cite{clow2025copsrobbersgraphspath,treewidth,gahlawat2023parameterizedanalysiscopsrobber}\nocite{treewidth}\nocite{gahlawat2023parameterizedanalysiscopsrobber}, offering evidence that classical ML is extracting meaningful structural information. This alignment provides confidence that future ML models can be trained using these features for cop number prediction.

\section{Discussion}\label{sec:disc}
\subsection{Limitations}
Several limitations to the study should be acknowledged. First, dataset bias poses a potential limitation. The dataset is dominated by graphs with cop number 1 and 2, while graphs with cop number 3 are comparatively rare. This is an inherent characteristic of small graph families. We used class weighting and stratified splits to reduce imbalance, however, the models may still be biased toward majority classes. Strong performance on higher cop numbers may reflect limited structural diversity rather than true generalization.

Second, feature correlation affects interpretability. Many graph invariants are highly correlated, while SHAP assumes feature independence. Importance may be distributed across correlated feature groups, and importance of a signle feature should be interpreted cautiously despite cross-checking with permutation importance.

Finally, external validity is limited by graph size and distribution. The study focuses on relatively small graphs and the expansion uses some degree restrictions, so generalization to larger graphs remains an open question.


\subsection{Future Work}
Future work could explore more advanced architectures such as graph foundation models, which are pretrained on more diverse graph datasets. Specialized attention mechanisms may capture generic structural patterns and could provide more transferable representations of graph embeddings. Fine-tuning pretrained graph encoders on cop number prediction may improve generalization on larger graph families. Additionally, incorporating highest ranked importance features learned in this study beyond degree may prove performance on network models.

\section{Conclusion}\label{sec:conc}
This work shows that the cop number of small graph families can be predicted accurately using both classical machine learning models and graph neural networks. Histogram-Gradient Boosting performs well using simple graph invariants, with interpretability analyses confirming reliance on structural properties consistent with known theory, such as connectivity, clique structure, and treewidth. Graph neural networks further demonstrate that cop number is strongly encoded in graph topology, even without explicit feature engineering. Overall, these results reveal the value of machine learning as an exploratory tool for identifying structural patterns in pursuit–evasion games and motivating ML applications that go beyond the limits of exact computation.

\bibliographystyle{named}
\bibliography{ijcai26}

@article{NOWAKOWSKI1983235,
title = {Vertex-to-vertex pursuit in a graph},
journal = {Discrete Mathematics},
volume = {43},
number = {2},
pages = {235-239},
year = {1983},
issn = {0012-365X},
doi = {https://doi.org/10.1016/0012-365X(83)90160-7},
url = {https://www.sciencedirect.com/science/article/pii/0012365X83901607},
author = {Richard Nowakowski and Peter Winkler}
}

@phdthesis{quilliot1978jeux,
  title={Jeux et pointes fixes sur les graphes},
  author={Quilliot, Alain},
  year={1978},
  school={Ph. D. Dissertation, Universit{\'e} de Paris VI}
}

@article{PETR202211,
title = {A faster algorithm for Cops and Robbers},
journal = {Discrete Applied Mathematics},
volume = {320},
pages = {11-14},
year = {2022},
issn = {0166-218X},
doi = {https://doi.org/10.1016/j.dam.2022.05.019},
url = {https://www.sciencedirect.com/science/article/pii/S0166218X22001780},
author = {Jan Petr and Julien Portier and Leo Versteegen}
}

@book{book,
  author = {A. Bonato and R.J. Nowakowski},
  year = {2011},
  title = {The Game of Cops and Robbers on Graphs},
  publisher = {American Mathematical Society}
}

@article{minon10,
    author = {William Baird and Andrew Beveridge and Macalester College and Anthony Bonato and Paolo Codenotti and Aaron Maurer and John McCauley and Silviya Valeva},
    title = {On the minimum order of k-cop win graphs},
    journal = {Contributions to Discrete Mathemtics},
    volume={9},
    number={1},
    pages={70-84},
    year={2012}
}

@misc{2011,
  author       = {Anton Afanassiev and Rafael Villarroel},
  title        = {k-FIXED COP NUMBER},
  year         = {2020},
  howpublished = {\url{https://github.com/Jabbath/Cop-Number/tree/master}},
  note         = {Accessed: 2025-08-12}
}

@inproceedings{Hagberg2008,
  author    = {Hagberg, Aric A. and Schult, Daniel A. and Swart, Pieter J.},
  title     = {Exploring network structure, dynamics, and function using {NetworkX}},
  booktitle = {Proceedings of the 7th Python in Science Conference ({SciPy2008})},
  pages     = {11--15},
  year      = {2008},
  editor    = {Varoquaux, Ga{\"e}l and Vaught, Travis and Millman, Jarrod},
  address   = {Pasadena, CA USA},
  url       = {http://conference.scipy.org/scipy2008/paper_pdfs/hagberg_exploring_network_structure.pdf}
}

@article{TURCOTTE202174,
title = {4-cop-win graphs have at least 19 vertices},
journal = {Discrete Applied Mathematics},
volume = {301},
pages = {74-98},
year = {2021},
issn = {0166-218X},
doi = {https://doi.org/10.1016/j.dam.2021.05.012},
url = {https://www.sciencedirect.com/science/article/pii/S0166218X21002018},
author = {Jérémie Turcotte and Samuel Yvon}
}

@article{treewidth,
author = {Bonato, Anthony and Clarke, Nancy and Finbow, Stephen and Fitzpatrick, Shannon and Messinger, Margaret-Ellen},
year = {2013},
month = {08},
pages = {},
title = {A note on bounds for the cop number using tree decompositions},
volume = {9},
journal = {Contributions to Discrete Mathematics},
doi = {10.55016/ojs/cdm.v9i2.62188}
}

@misc{sun2025graphproptraininggraphfoundation,
      title={GraphProp: Training the Graph Foundation Models using Graph Properties}, 
      author={Ziheng Sun and Qi Feng and Lehao Lin and Chris Ding and Jicong Fan},
      year={2025},
      eprint={2508.04594},
      archivePrefix={arXiv},
      primaryClass={cs.LG},
      url={https://arxiv.org/abs/2508.04594}, 
}

@misc{ying2019gnnexplainergeneratingexplanationsgraph,
      title={GNNExplainer: Generating Explanations for Graph Neural Networks}, 
      author={Rex Ying and Dylan Bourgeois and Jiaxuan You and Marinka Zitnik and Jure Leskovec},
      year={2019},
      eprint={1903.03894},
      archivePrefix={arXiv},
      primaryClass={cs.LG},
      url={https://arxiv.org/abs/1903.03894}, 
}

@misc{sklearn-tree-doc,
  title        = {Tree-based Models — scikit-learn 1.5.0 documentation},
  author       = {Scikit-learn Developers},
  howpublished = {\url{https://scikit-learn.org/stable/modules/tree.html}},
}

@misc{sklearn-hgbt-example,
  title = {Features in Histogram Gradient Boosting Trees},
  author = {Scikit-learn Developers},
  howpublished = {\url{https://scikit-learn.org/stable/auto_examples/ensemble/plot_hgbt_regression.html}},
}

@manual{sklearn-logisticregression,
  title        = {sklearn.linear-model.LogisticRegression — scikit-learn},
  author       = {{scikit-learn developers}},
  year         = {2025},
  note         = {\url{https://scikit-learn.org/stable/modules/generated/sklearn.linear_model.LogisticRegression.html}},
}

@misc{sklearn-random-forest-doc,
  title        = {sklearn.ensemble.RandomForestClassifier — scikit-learn documentation},
  author       = {Scikit-learn Developers},
  howpublished = {\url{https://scikit-learn.org/stable/modules/generated/sklearn.ensemble.RandomForestClassifier.html}},
}

@misc{douglas2011weisfeilerlehmanmethodgraphisomorphism,
      title={The Weisfeiler-Lehman Method and Graph Isomorphism Testing}, 
      author={B. L. Douglas},
      year={2011},
      eprint={1101.5211},
      archivePrefix={arXiv},
      primaryClass={math.CO},
      url={https://arxiv.org/abs/1101.5211}, 
}

@article{BonatoChiniforooshanPralat2010,
  title        = {Cops and Robbers from a distance},
  author       = {Bonato, Anthony and Chiniforooshan, Ehsan and Pra{\l}at, Pawe{\l}},
  journal      = {Theoretical Computer Science},
  volume       = {411},
  number       = {43},
  pages        = {3834--3844},
  year         = {2010},
  doi          = {10.1016/j.tcs.2010.07.003},
  url          = {https://math.ryerson.ca/~abonato/papers/TCS7932.pdf}
}

@article{KINNERSLEY2015201,
title = {Cops and Robbers is EXPTIME-complete},
journal = {Journal of Combinatorial Theory, Series B},
volume = {111},
pages = {201-220},
year = {2015},
issn = {0095-8956},
doi = {https://doi.org/10.1016/j.jctb.2014.11.002},
url = {https://www.sciencedirect.com/science/article/pii/S0095895614001282},
author = {William B. Kinnersley}
}

@manual{pytorch_dropout,
  title        = {PyTorch {\tt torch.nn.Dropout} Documentation},
  author       = {{PyTorch Contributors}},
  organization = {PyTorch},
  year         = {2025},
  url          = {https://docs.pytorch.org/docs/stable/generated/torch.nn.Dropout.html}
}

@article{https://doi.org/10.1002/jgt.23194,
author = {Kenter, Franklin and Meger, Erin and Turcotte, Jérémie},
title = {Improved bounds on the cop number when forbidding a minor},
journal = {Journal of Graph Theory},
volume = {108},
number = {3},
pages = {620-646},
doi = {https://doi.org/10.1002/jgt.23194},
url = {https://onlinelibrary.wiley.com/doi/abs/10.1002/jgt.23194},
eprint = {https://onlinelibrary.wiley.com/doi/pdf/10.1002/jgt.23194},
year = {2025}
}

@misc{bonato2017characterizationsalgorithmsgeneralizedcops,
      title={Characterizations and algorithms for generalized Cops and Robbers games}, 
      author={Anthony Bonato and Gary MacGillivray},
      year={2017},
      eprint={1704.05655},
      archivePrefix={arXiv},
      primaryClass={math.CO},
      url={https://arxiv.org/abs/1704.05655}, 
}

@article{Mohar_2025, 
title={The game of Cops and Robber on geodesic spaces}, volume={77}, DOI={10.4153/S0008414X24000543}, number={6}, journal={Canadian Journal of Mathematics}, author={Mohar, Bojan}, year={2025}, pages={1827–1860}
}

@misc{gahlawat2023parameterizedanalysiscopsrobber,
      title={Parameterized Analysis of the Cops and Robber Problem}, 
      author={Harmender Gahlawat and Meirav Zehavi},
      year={2023},
      eprint={2307.04594},
      archivePrefix={arXiv},
      primaryClass={cs.DM},
      url={https://arxiv.org/abs/2307.04594}, 
}

@book{diestel2017graph,
  title     = {Graph Theory},
  author    = {Diestel, Reinhard},
  edition   = {5},
  publisher = {Springer},
  year      = {2017},
  series    = {Graduate Texts in Mathematics},
  volume    = {173}
}

@misc{clow2025copsrobberscliquecovers,
      title={Cops and Robbers, Clique Covers, and Induced Cycles}, 
      author={Alexander Clow and Imed Zaguia},
      year={2025},
      eprint={2507.14321},
      archivePrefix={arXiv},
      primaryClass={math.CO},
      url={https://arxiv.org/abs/2507.14321}, 
}

@article{HOANG1987302,
title = {On a conjecture of Meyniel},
journal = {Journal of Combinatorial Theory, Series B},
volume = {42},
number = {3},
pages = {302-312},
year = {1987},
issn = {0095-8956},
doi = {https://doi.org/10.1016/0095-8956(87)90047-5},
url = {https://www.sciencedirect.com/science/article/pii/0095895687900475},
author = {C.T Hoàng}
}

@misc{baird2013meynielsconjecturecopnumber,
      title={Meyniel's conjecture on the cop number: a survey}, 
      author={William Baird and Anthony Bonato},
      year={2013},
      eprint={1308.3385},
      archivePrefix={arXiv},
      primaryClass={math.CO},
      url={https://arxiv.org/abs/1308.3385}, 
}

@misc{bonato2022meynielextremalfamiliesgraphs,
      title={On Meyniel extremal families of graphs}, 
      author={Anthony Bonato and Ryan Cushman and Trent G. Marbach},
      year={2022},
      eprint={2201.08719},
      archivePrefix={arXiv},
      primaryClass={math.CO},
      url={https://arxiv.org/abs/2201.08719}, 
}

@misc{clow2025copsrobbersgraphspath,
      title={Cops and Robbers on Graphs with Path Constraints}, 
      author={Alexander Clow and Erin Meger},
      year={2025},
      eprint={2509.10941},
      archivePrefix={arXiv},
      primaryClass={math.CO},
      url={https://arxiv.org/abs/2509.10941}, 
}

@article{Kosowski2015,
  author    = {Kosowski, Adrian and Li, Bin and Nisse, Nicolas and Suchan, Karol},
  title     = {$k$-Chordal Graphs: From Cops and Robber to Compact Routing via Treewidth},
  journal   = {Algorithmica},
  volume    = {72},
  number    = {3},
  pages     = {758--777},
  year      = {2015},
  doi       = {10.1007/s00453-014-9871-y},
  url       = {https://doi.org/10.1007/s00453-014-9871-y},
  issn      = {1432-0541}
}

\end{document}